\documentclass[twoside,11pt]{article}

\usepackage[abbrvbib, preprint]{jmlr2e}

\def\CC{{C\nolinebreak[4]\hspace{-.05em}\raisebox{.4ex}{\tiny\bf ++}}}

\jmlrheading{0}{0000}{00-00}{0/00}{00/00}{olsson00a}{Roland Olsson, Chau Tran and Lars Magnusson}

\ShortHeadings{Automatic Neuron Synthesis}{Olsson, Tran and Magnusson}
\firstpageno{1}

\begin{document}

\title{Automatic Synthesis of Neurons for Recurrent Neural Nets}

\author{\name Roland Olsson \email Roland.Olsson@hiof.no \\
  \name Chau Tran \email Chau.Tran@hiof.no \\
  \name Lars Magnusson \email Lars.Magnusson@hiof.no \\
  \addr Department of Computer Science\\
  {\O}stfold University College\\
    N-1757 Halden, Norway
  }

\editor{Nomen Nescio}

\maketitle

\begin{abstract}

We present a new class of neurons, ARNs, which give a cross entropy
on test data that is up to three times lower than the one achieved by
carefully optimized LSTM neurons.
The explanations for the huge improvements that often are achieved are
elaborate skip connections through time, up to four
internal memory states per neuron and a number of novel
activation functions including  small quadratic forms.

The new neurons were generated using automatic programming
and are formulated as pure functional programs that easily
can be transformed.

We present experimental results for eight datasets and found
excellent improvements for seven of them, but LSTM
remained the best for one dataset.
The results are so promising that
automatic programming to generate
new neurons should become part of the standard operating procedure
for any machine learning practitioner who works on time series data
such as sensor signals.

\end{abstract}


\section{Introduction}

Time series or sequence data is abundant in machine learning and range from signal processing
to stock prices and more complex data such as natural language. Indeed, the entire life of a human being is also
a  time series dataset generated by the five senses.

This paper focuses on improving LSTM neurons \citep{hochreiter1997long}
since they both are theoretically interesting and also
one of the most popular and effective machine learning tools.

By using the ADATE automatic programming system \citep{olsson1995inductive},
we generated new neurons for each specific dataset and found
that they were up to three times better than LSTM as measured
by cross entropy on test data. We used eight datasets with between 22 and 500 timesteps and found
the biggest improvement for the 500 timestep dataset.

The main novelties of the ADATE Recursive Neurons (ARNs) are as follows.

\begin{enumerate}

\item ARNs often contain  skip connections like in  Highway nets and ResNet, 
but through the time dimension.

\item Up to four memory states per neuron instead of just one as in the LSTM.

\item Novel activation functions built from, for example, small quadratic forms and sequences of a
set of three predefined and  well known activation functions.

\item A given neuron may use linear combinations of the states of other neurons.

\item Differentiating between recurrent output as well as state values from ``self'' and ``all others''.

\end{enumerate}

We will now give an overview of the paper.
First, we present related work on automatic programming and evolution of recurrent neurons.
The next section describes how to represent ARNs in a small and
purely functional subset of the Standard  Meta Language (SML) \citep{milner1997definition} 
and also gives a compact and purely functional definition of
the classic LSTM as an illustration. We then provide experimental results for eight medium size datasets
that typically are found in practical sensor data analysis applications. 
We finish by some conclusions and a glimpse of the almost infinite possibilities for future work that
are enabled by the technology in this paper.
The appendix contains seven examples of ARNs to exhibit some of their novelties.

\section{Related work}

Long short-term memory (LSTM) was originally presented in Sepp Hochreiter's masters thesis 
\citep{hochreiter1997long} with J{\"u}rgen Schmidhuber as the advisor
and
made more well known by Hochreiter and Schmidhuber \citep{hochreiter1997long}.

It remains a quite popular and competitive technology  also 25 years later, which
is remarkable in a rapidly developing field such as machine learning.
We assume that the reader already is familiar with LSTMs and will not describe them here.

Some newer and prominent recurrent neurons that sometimes outperform LSTM include
Gated Recurrent Units \citep{cho2014properties}
and Independently Recurrent Neural Nets \citep{li2018independently}.

\subsection{Using evolutionary computation to synthesize recurrent neurons}

Galvan and Mooney \citep{galvan2021neuroevolution} provide an extensive review
of neuroevolution and show that the field primarily deals
with using evolutionary methods for hyperparameter or architecture and topology search for entire
neural nets using a predefined set of neuron types.
They do not mention any work that evolves the neurons themselves.

An early attempt to evolve neurons using ADATE was made by Berg et.\ al.\ \citep{berg2008automatic}, but these neurons
were a special type of spiking neurons for segmentation of gray scale images
containing noisy rectangles. Although reasonable improvements were found, the results are not practically competitive
and have been superseded by modern CNN based nets for image processing.

A key distinction between different approaches to neuron evolution is whether
the goal is to find a general neuron that is better for a large class of datasets
or whether it is to produce more specialized neurons that  work best for an application
specific class of datasets. Our goal is  to produce superior and application specific neurons
whereas Bayer et.\ al\, \citep{bayer2009evolving} and Jozefowicz et.\ al.\ \citep{jozefowicz2015empirical} try to produce more general
replacements for the LSTM, which may be much more difficult to achieve.

Of course, the disadvantage with application specific neurons, for example neurons specialized for high frequency trading,
is that it is more difficult to replace the neuron itself in the well known neural toolboxes than it is to just
optimize the hyperparameters and the architecture. A possible advantage though is that much bigger application
specific improvements may be found. Thus, we argue that all three of overall architecture, hyperparameters and neuron design need
to be optimized for the best results.

Bayer et.\ al\, \citep{bayer2009evolving} mention another distinction, namely if a very indirect representation such as DNA
is  employed to represent neurons or whether transformations such as so-called ``mutations'' are performed 
directly on the neuron circuits. They use the latter approach and  define a number of well designed 
transformations on a directed acyclic graph 
representation of neurons. Thus, they build an evolutionary system from scratch expressly for evolving LSTM
replacements. Our approach is a bit different since we use the universal ADATE automatic programming system
and do not code any evolutionary search nor any transformations specifically for neuron evolution.

Jozefowicz et.\ al.\ \citep{jozefowicz2015empirical} also build their evolutionary computation system from scratch
and excel in neural architecture design and  hyperparameter search. It is obvious from the paper
that they have deep expertise in these areas.
In contrast to us, they perform
hyperparameter search  on-the-fly during the evolution. 
As a result of this and other experimental choices, they  evaluate a number of new neuron candidates 
that is about five orders of magnitude smaller than we did with ADATE.

\subsection{Automatic design of algorithms through evolution}

Automatic design of algorithms through evolution (ADATE) \citep{olsson1995inductive} was originally developed
for automatic synthesis of symbolic and recursive functional programs, but a few years later
also given simple mechanisms for optimizing floating point constants.
It writes programs in a very small subset of SML \citep{milner1997definition}, which was conceived by Robin Milner
as a meta language for symbolic logic \citep{gordon1978metalanguage}. As we will see in this paper, it is also quite suitable
for meta machine learning.

Some other and fundamentally different approaches to inductive programming,
that are more suitable for explainable machine learning than ADATE, are 
discussed in the JMLR special issue edited  by \citep{ kitzelmann2006inductive}.

ADATE has the ability to invent help functions as they are needed and to perform so-called  lifting and
distribution transformations on case- and let-expressions as well as many other semantics preserving
and useful program transformations \citep{olsson1995inductive}. However, the details of these are beyond the scope of the current paper.

ADATE needs a specification of the problem to be solved.
A specification contains the following primary ingredients.

\begin{enumerate}

\item Algebraic datatypes, for examples various lists and trees.

\item A signature for the function {\tt f} that is to be synthesized and optionally also an initial
definition of that function.

\item An evaluation function that tests a synthesized function on a number of training inputs.
Note that {\tt f} may occur as a small part of a big program and that
calculating {\tt f} typically means to run this program on the training inputs.
Thus, ADATE performs symbolic reinforcement (meta-)learning.

\end{enumerate}

Given the specification, ADATE runs on a cluster where many programs are transformed and evaluated in parallel.
The result is a Pareto front of gradually more syntactically complex and better programs.
Syntactic complexity is the sum of the base-2 logarithm of the occurrence probabilities
of the symbols in all nodes in the syntax tree of a program.
It falls on the human operator of ADATE to select a suitable program from the Pareto front.
Below, we will make it simple and always choose the one that has the lowest mean squared error or cross entropy
on validation data.

\section{How to represent ARNs as functional programs}

We will first explain how to represent neurons and then explain how a neuron
is a transition function from one timestep to the next. Then, we will give
a functional program corresponding to an LSTM neuron as an example.

\subsection{Datatypes}

A key ingredient in neural nets is linear combination of a set of 
floating point values and we will start by explaining how to  represent this
using algebraic datatypes. For example, consider a linear combination of
two values $x_1$ and $x_2$ using weights $w_1$ and $w_2$ and bias $b$.

\[ 
w_1 x_1 + w_2 x_2 + b
\]

In our algebraic datatype, we abstract away the weights and represent a linear combination
using the SML type {\tt linComb}, defined below, with two constructors {\tt bias} and {\tt cons.} 
The {\tt *} is Cartesian product and the {\tt |} introduces a sum type.

\begin{verbatim}
datatype linComb = bias | cons of real * linComb
\end{verbatim}

This is essentially the type of lists that always contain at least one value, namely a bias.
The type {\tt real} represents floating point numbers in spite of its SML name.
The linear combination above would correspond to the following list.

\begin{verbatim}
cons( x1, cons( x2, bias ) )
\end{verbatim}

At this point, the reader perhaps wonder how on earth the weights will be specified.
The answer is that we only allow a limited number of weight mappings, where each mapping
is indicated by a function called {\tt lc}$i$ for $i = 0, 1, \ldots, n$ for $n+1$ mappings.
The current implementation  has $n=4$, which means that there will be five different sets
of forward, recurrent and peep weight matrices.

The expression below will be translated by our compiler to
the linear  combination above  assuming that mapping 0 has weights $w_1$ and $w_2$ and bias $b$.

\begin{verbatim}
lc0( cons( x1, cons( x2, bias ) ) )
\end{verbatim}

However, if we were to use say {\tt lc1} instead of {\tt lc0}, another bias and set of weights
would be used after compilation.

\subsection{Transition function}

It is now time to start explaining how to represent recursive neural net neurons and
their internal and external connections.

To introduce our notation, let us first consider the transition function $f$  
of type $\mathbb{R}^2 \rightarrow \mathbb{R} $ for a simple RNN neuron with 
input $ x^{(t)} $ and output $ y^{(t)} $ at time $t$.
Thus, the inputs and outputs of $f$ are as follows.

\[
f( x^{(t)}, y^{(t)} ) = y^{(t+1)}
\]

An LSTM neuron has an internal state $s$ which means that 
its type is  $\mathbb{R}^3 \rightarrow \mathbb{R}^2 $. 

\[
f( x^{(t)}, s^{(t)}, y^{(t)} ) = ( s^{(t+1)}, y^{(t+1)} )
\]

An ADATE recursive neuron (ARN) has four internal states instead of just one which
implies that it could be a function  with the following signature.

\[
f( x^{(t)}, s_0^{(t)}, s_1^{(t)}, s_2^{(t)}, s_3^{(t)}, y^{(t)} ) =
 ( s_0^{(t+1)}, s_1^{(t+1)}, s_2^{(t+1)}, s_3^{(t+1)}, y^{(t+1)} )
\]

However, we will differentiate between the output from the neuron itself at timestep $t$ and
the linear combination of the outputs of the other neurons in the same layer.
This ``recognition of self'' turns out to be an  important feature of ARNs as shown by the 
experimental results in Section~{\ref{ExperimentalResults}}.

Experimentally, we also found that skip connections through time were invented by ADATE and it is easy to see that
this is enabled by having many internal states instead of just one.
For example, if we wish the output at timestep $t$ to be available as an input at timestep $t+2$,
we can choose $s_1^{(t+1)}$ to $y^{(t)}$ and $s_2^{(t+1)}$ to $s_1^{(t)}$. 
Many other and much more intricate ways of curating information through time are possible and
have been invented by ADATE.

\subsection{Functional programming definition of an ARN}

In order to explain the signature of an ARN, we will switch to functional programming notation
and start by explaining the input parameters.

\begin{description}

\item[{\tt InputsLC : linComb.}] The list of inputs. 

\item[{\tt SelfPeep0 : real.}] State $s_0^{(t)}$.
\item[{\tt SelfPeep1 : real.}] State $s_1^{(t)}$.
\item[{\tt SelfPeep2 : real.}] State $s_2^{(t)}$.
\item[{\tt SelfPeep3 : real.}] State $s_3^{(t)}$.

\item[{\tt SelfOutput : real.}] The current output value $y_{t}$.
\item[{\tt OtherPeepsLC : linComb.}] The list of the $s_0^{(t)}$ values of all other nodes in the layer.
\item[{\tt OtherOutputsLC : linComb.}]  The list of the  $y^{(t)}$  values of all other nodes in the layer.

\end{description}

Finally, we  will explain how variables are bound to values of expressions in
the small subset of SML used in ADATE.
The syntax is a bit unusual and relies on case-expressions for this. For example,
the  case-expression below means that the variable {\tt V} will represent the 
value of the expression {\tt E1} and that its scope is {\tt E2}. 

\begin{verbatim}
case E1 of V => E2
\end{verbatim}

Of course, these case expressions are used to construct functional programs
that correspond to directed acyclic graphs (DAGs) in neural circuit diagrams.

Consider the usual LSTM neuron with peepholes in Figure~\ref{LSTMFunctional} that is a direct translation of
the LSTM circuit diagram in \citep{greff2016lstm}.
We now have everything that is needed to understand the functional programming
definition of this neuron.
Figure~\ref{LSTMFunctional} contains five case-expressions
before giving the output tuple in which only the first and the last fields
are used and the three in the middle just contain dummy values,
in this case $0.0$, since they are not needed for LSTM.
Recall that the output quintuple first contains
the four next states and then the next output as its last field.
Since all our datasets are centered and scaled, it turns out that biases are almost totally redundant
in the LSTM, which means that they have been omitted in Figure~\ref{LSTMFunctional}.

\begin{figure}
\begin{verbatim}
    case
      tanh( 
        lc0 InputsLC + lc0( cons( SelfOutput, OtherOutputsLC ) )
        )     
    of Z =>
    case
      sigmoid(
        lc1 InputsLC +
        lc1( cons( SelfPeep0, cons( SelfOutput, OtherOutputsLC ) ) )
        )     
    of I =>
    case
      sigmoid(
        1.0 + lc2 InputsLC +
        lc2( cons( SelfPeep0, cons( SelfOutput, OtherOutputsLC ) ) )
        )     
    of F =>
    case F * SelfPeep0  +  I * Z  of S0 =>
    case
      sigmoid(
        lc3 InputsLC +
        lc3( cons( S0, cons( SelfOutput, OtherOutputsLC ) ) )
        )     
    of O =>
      ( S0, 0.0, 0.0, 0.0, O * tanh S0 )
\end{verbatim}

\caption{The LSTM neuron with peepholes as a functional program.}
\label{LSTMFunctional}

\end{figure}

As another example, we will explain a rather simple neuron syntesized by ADATE
and that comes from the Pareto front for Double pendulum discussed later in Section~\ref{ParetoFront}.

The program corresponding to the neuron is given in Figure~\ref{PendulumProgram}.
The auxiliary function {\tt g} has been invented by ADATE.
Its purpose is to bind {\tt S0} to the {\tt tanh} expression.
The return value of {\tt g} is a quintuple where the last field is
a  quadratic polynomial that gives the output from the neuron.
The first field is just {\tt S0},  which will become {\tt SelfPeep0} for
the next timestep. A linear combination of these values for
the other neurons than ``self'' will become {\tt OtherPeepsLC}.

\begin{figure}
\begin{verbatim}
let
  fun g S0 = ( S0, S0, S0, S0, S0 - S0 * S0 )
in
  g(
    tanh(
      relu( lc2 InputsLC + SelfPeep0 ) -
      lc1( cons( lc0 OtherPeepsLC, InputsLC ) )
      ) )
end
\end{verbatim}
\caption{A small neuron for modeling a double pendulum.}
\label{PendulumProgram}
\end{figure}

\begin{figure}[!htbp]
\begin{center}
\scalebox{0.40}
 {
   {\includegraphics{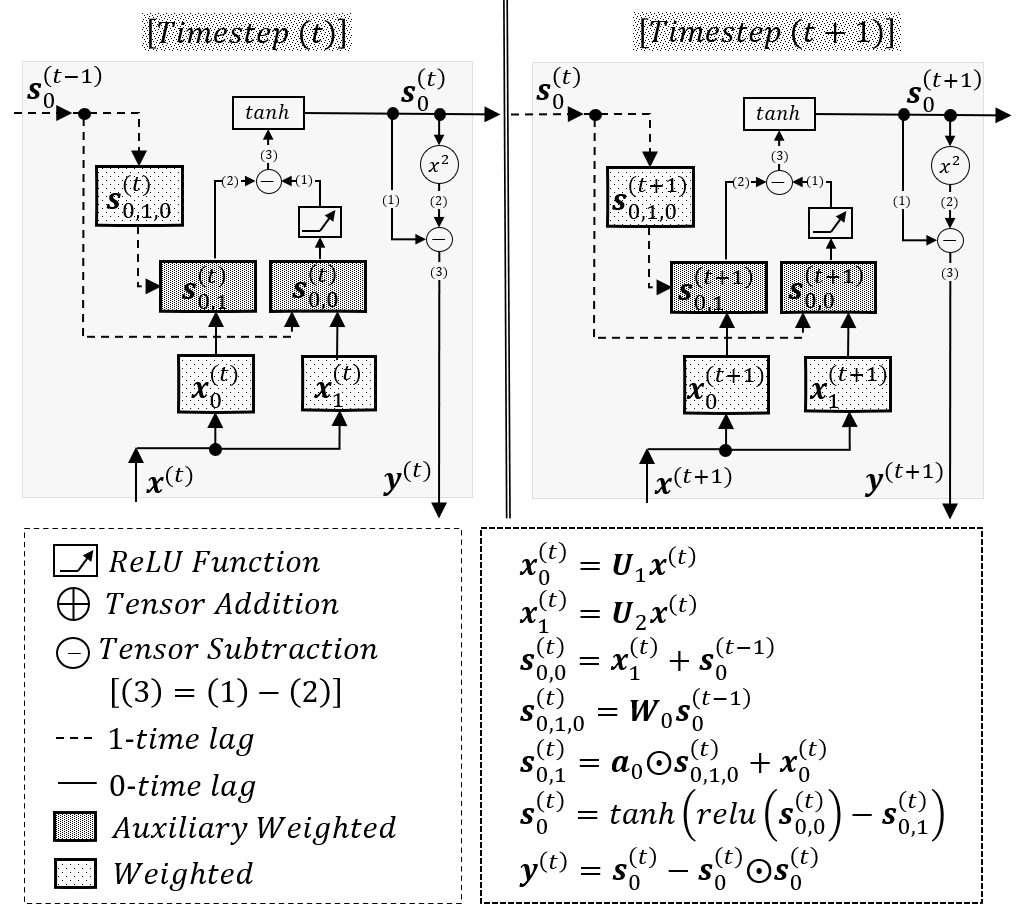}}
 }
\end{center}
\caption{A circuit diagram for the neuron in Figure~\ref{PendulumProgram}.}
\label{PendulumFig}
\end{figure}

Figure~\ref{PendulumFig} shows a more traditional representation of the neuron
with one copy of the neuron for timestep $t$ and another for $t+1$
with dashed lines indicating information flow across timesteps.
The figure uses linear algebra notation with bold lower case variables being vectors, whereas
bold upper case variables represent matrices.  The vector of state 0 values is ${\bf s_0}$, corresponding to 
the scalar variable {\tt S0} in the program above. The input weight matrices ${\bf U_1}$ and ${\bf U_2}$
correspond to the weight mappings indicated by {\tt lc1} and {\tt lc2} respectively.
The recurrent weight matrix ${\bf W_0}$ is hollow, that is has zeroes along its diagonal to
exclude ``self'' from {\tt OtherPeepsLC}.
The vector ${\bf a_0}$ contains so-called auxiliary weights and is multiplied
element-wise (Hadamard product)  as indicated by the operator $\odot$. 
It is not really needed here but corresponds to  how the {\tt cons} is translated
by our compiler.

The shaded boxes in the diagram implicitly contain weight vectors and matrices, which have been omitted
in order not to clutter the diagram.

The intermediate vector ${\bf s_{0,0} }$ represents    {\tt lc2 InputsLC + SelfPeep0}
and
${\bf s_{0,1} }$ represents      {\tt lc1( cons( lc0 OtherPeepsLC, InputsLC ) ).}
When we introduce new intermediate variables, we use a second subscript as above
to differentiate them

In our actual {\CC } code, all input weight matrices are stacked in one matrix and
the same goes for all recurrent weight matrices in order to reduce
the number of matrix-vector multiplications.

\subsection{Choosing activation functions}

ADATE needs to be given a set of predefined functions to use for program synthesis.
Obviously, we include addition, subtraction, multiplication and division.
It would be possible for ADATE to use these together with if-expressions
to synthesize its own activation functions, for example ReLU \citep{fukushima1969visual}
or good approximations of the hyperbolic ones, 
but after some preliminary experimentation we
decided against using if-expressions altogether.
 
Instead, we chose to include three predefined activation functions, which turn out to be 
very useful for synthesizing other and novel activation functions.

First, we chose to include {\tt tanh} but not sigmoid since ADATE easily can express it as
a scaled version of {\tt tanh}.

Second, the {\tt relu} function was included and after that we also chose
to add a linear approximation of {\tt tanh} which we called saturated ReLU ({\tt srelu}).
The latter is defined as being the identity function for values between -1 and 1,
just 1 for values greater than 1 and -1 for values less than -1.
It could be implemented using two ReLU functions, but we chose to relieve ADATE of that extra burden.

Our intuition was that elimination of the more curved parts of {\tt tanh} may lead
to smaller problems with vanishing gradients since {\tt relu} also is linear and seems
to have this advantage. However, we still wanted a function more suitable for gating
than {\tt relu} and therefore defined {\tt srelu}.

Thus, our set of predefined activation functions is {\tt tanh}, {\tt relu} and {\tt srelu}.

\section{Experiments}

In this section, we will first explain our neural net architecture and hyperparameter optimization, 
then how to very quickly evaluate neuron candidates
and then introduce the datasets that we tested ARN on. Finally, we will present
the experimental results along with statistical analyses.

In general, we always split datasets into three subsets, with 50\% for training, 25\% for validation
and 25\% for testing. The test set is used only once at the very end to prevent subtle forms of
information leaks and the overfitting that may result.

\subsection{Architecture and hyperparameter optimization}

Before starting an ADATE run for a new dataset, we optimized the architecture and the hyperparameters for LSTM
and used these also for the neurons discovered by ADATE. 
Assume that we have a dataset with $n_o$ outputs and that there  are $l$ nodes
in the LSTM layer.

We used the simple net architecture below.

\begin{enumerate}

\item An LSTM layer with $l$ nodes.

\item A hyperbolic tangent layer with $n_o$ nodes.

\item A linear layer with $n_o$ nodes.

\end{enumerate}

We use Glorot initialization for forward weights and orthogonal initialization
for recurrent weights with a scaling factor of 0.1 in both cases. Also, we use an initial extra bias
of 1 for the forget gates as recommended by \citep{jozefowicz2015empirical} and others.

The loss function was mean squared error (MSE) for regression and softmax plus cross entropy for classification.
We tried a variety of other backends such as one or more ReLU layers instead of the hyperbolic tangent layer,
but found no significant  gains in validation losses.
Before doing any other optimization, we found the optimal number $l$ of LSTM nodes under the assumption 
that $l = 2^i$ for $i=1,2, \ldots, 7$. The reason that we could use a rather small upper bound for the number of nodes
is that we did not use datasets such as those from NLP that require a big memorization ability.

The next topic is hyperparameter optimization.
Choi et.\ al.\ \citep{choi2019empirical} found that comparisons of different weight optimizers
such as ADAM \citep{kingma2014adam} and NADAM \citep{dozat2016incorporating} often is skewed by inadequate hyperparameter
optimization and recommend that all parameters in the algorithms should
be optimized. We used their methodology with the initial parameter ranges that they give in Appendix~D.8
for every single dataset. 
As they recommend, we performed a random search within these ranges but used many more evaluations, 
namely 512 random parameter sets for each dataset and
picked the best one for each dataset using the validation data.
Thus, we optimized all parameters in the ADAM algorithm and not just the learning rate.

We tried with both ADAM and NADAM, but the difference between them was negligible. ADAM was used
in all subsequent experiments.

We also used the same learning rate schedule as Choi et.\ al.\ \citep{choi2019empirical}, namely the one 
from  Shallue et.\ al.\ \citep{shallue2019measuring}, and
optimized the initial learning rate, the duration of the decay and the amount of decay together with the ADAM hyperparameters.
Our batch size is so small that no warm-up is needed.

The amount of training is often measured in the number of epochs, but we will instead use the total number
of training examples including duplicates, that is the number of epochs times the training set cardinality.
Each training session consisted of running 320 000 training examples with a batch size
of four, that is a total of 80 000 weight updates.
After preliminary experiments, we abandoned L1 and L2 regularization and dropout but chose to employ
model checkpointing as follows. For every 20 000 training examples, we ran
the net with the current weights on the validation data and updated the best weights found if there was
a validation improvement.

\subsection{Experiment design and implementation}

A key challenge with ADATE as for many other evolutionary algorithms is that very many
evaluations may be required, where evaluation means to compute an evaluation value, sometimes called ``fitness'',
for a solution candidate, which here is an ARN.

ADATE first screens candidates using a quite limited and very quick evaluation.
The candidates that pass the screening, typically less than 1\%, then move
on to a second stage evaluation that in most cases use at least 100 times more time
per candidate. Finally, less than 1\% of the ones who pass the second stage
are then passed to a third and final stage. The first two stages use a smaller number
of training steps and a smaller number of  recurrent nodes as follows.

\begin{description}

\item[First stage.]
The goal for this stage was to have an overall evaluation time for a new neuron that is less than 100 ms.
For some CPUs and datasets, we achieved 50 ms and for other combinations it is up to 200 ms.
Of course, the evaluation results will be very bad in this stage, but still enough to filter away really bad
programs. The number of nodes was always four and we used only 5000 training examples.
Additionally, we restricted the training to run on only the five last timesteps in each time series.

\item[Second stage.]
Here, we use 40 000 training examples, all timesteps and $l/4$ nodes.
In comparison with the first stage, the run time is increased by a factor given by
the following expression where $n_t$ is the number of time steps.

\[
8 \cdot (l/16)^2 \cdot n_t / 5
\]

For example, with $l=64$ and 100 time steps, the increase is 2560 times.

\item[Third stage.]
This stage uses the full 320 000 training examples and $l$ nodes.
Thus, its expected run time is $8 \cdot 4^2$, that is 128,  times longer than the second  stage.

\end{description}

For say $l=64$ and 100 timesteps, the total speed amplification is about $2560 \cdot$$ 128 /$$ 3$, 
that is about $10^5$ times, since
the implementaion actually tries to spend equally long run time on each stage.

Our  compiler translates a functional program that represents a neuron into byte code, which
then is run on-the-fly by a byte code interpreter implemented in \CC. Since matrix operations
require almost all the execution time, it does not matter that neuron evaluation is somewhat slower
than optimized machine code.

All of the neural net is also implemented in {\CC } and uses the automatic differentiation library AADC
that kindly was  made available by Matlogica Ltd, who also provided absolutely outstanding support
for how to use it in the most efficient way, especially for matrix operations. The resulting vectorized machine code,
that is AVX256 or optionally AVX512, 
appears to run several times faster than Tensorflow, when the latter is restricted to a single CPU core.
We evaluated the C interfaces of all the leading neural net packages but did not find anything that
could compete with the speed, stability and ease of use that characterizes AADC.

To check the correctness of our implementation, we ran the same datasets with LSTM  using both our {\CC } code and Tensorflow
and compared the results.  Additionally, we implemented one of the new neurons in Tensorflow and compared
with the {\CC } code without finding any  statistically significant differences. All comparisons and other experiments used
64-bit floating point numbers since a 32-bit  AADC version was not yet available from Matlogica.

An ADATE run used between 500 and 1000 CPU cores depending on what was available. The number of evaluated
candidate neurons was on the order of one billion for each dataset and around three days of run time
were typically needed to find the best results.

\subsection{Datasets}

Since we need to evaluate new neurons rather quickly, we chose to focus on datasets
that can be effectively run with 128 LSTM neurons or less, that is rather small nets.
For this reason, we avoided speech and NLP datasets that typically require bigger nets as 
well as attention \citep{bahdanau2014neural,graves2013generating} to get the best results.

Many of the datasets come from the excellent online repository for time series datasets
that is curated by Bagnall et.\ al.\ \citep{bagnall16bakeoff} and their time series classification (TSC)
research group. Most of the datasets
are time series from various physical sensors.

The only preprocessing that we employed was either centering and scaling or one hot encoding, depending
on whether a predictor or an output was ordinal or nominal. Note that  the goal of the experiments
is a reliable comparison between LSTM and new neurons and not to get state-of-the-art
results for any dataset. For this reason, we did not do any feature engineering or data augmentation.
Thus, due to not doing any more advanced preprocessing, our results may for some datasets
be far from the best that is possible.

We will first give a brief description of each dataset and then
list the number of examples, that is time series, and the number of predictors for each one.

\begin{description}

\item[3W]
This is a predictive maintenance dataset that was donated to the UCI machine learning repository 
by Petrobras \citep{vargas2019realistic}
and  consists of actual sensor values and events in oil wells supplemented by simulated and handcrafted
values and events.
The task is to predict undesirable events.
We preprocessed the dataset into non-overlapping time windows and used only sensor
readings, for example temperature and pressure, as  predictors.

\item[Crop]
This dataset is from TSC but was originally collected by Tan et.\ al.\ \citep{tan2017indexing}.
The goal is to classify the type of crop
on a patch of land from the time series of images produced by the Sentinel-2A satellite, which
has plant growth monitoring as one of its primary missions.

\item[Double pendulum]
The task of the RNN is to learn to simulate a double pendulum.
In general, RNNs can be applied to modeling of chemical and physical processes and are, for example,
used quite successfully for this purpose by the Borregaard biorefinery since the beginning of 2022.
Inspired by this, we generated the Double Pendulum dataset from the simulation provided with the SimBody physics engine.
The predictors are the center of gravity coordinates of the first as well as the second pendulum at a given timestep
and the response is the corresponding coordinates for the next timestep. 
Each time series is generated from randomly chosen initial positions and angular velocities.
We deliberately lowered the sampling frequency so that black box modeling from measurement data alone becomes challenging.

\item[ECG5000]
This dataset is a 20-hour long ECG and was originally published by Goldberger et.\ al.\ \citep{goldberger2000physiobank},
who preprocessed the signals to make each heartbeat equally long using interpolation. We downloaded it from TSC.

\item[FordB]
This dataset was originally used in a competition at WCCI 2008 \citep{nasr2008} and 
the task is to determine if there is something wrong with an internal combustion engine
based on recordings of the engine sound.  This dataset was also taken from TSC.

\item[Insect wingbeat]
This dataset \citep{chen2014flying} contains spectrograms from the sounds generated by the wings of insects, in this case
mostly mosquitoes and flies. The goal is to classify the species of an insect based on a sound recording.
We downloaded it from TSC.

\item[LSST]
This is yet another TSC dataset but originally comes from a 2018 Kaggle competition \citep{allam2018photometric}.
It contains simulated data from the Vera C. Rubin Observatory, also known as  the Large Synoptic Survey Telescope (LSST),
that are measurements of how the brightness of an  astronomical object varies with time.
The goal is to determine which kind of object that was observed.

\item[WISDM]
The task is to predict the activity of a cell phone user based on
the time series of accelerometer values \citep{kwapisz2011activity}.
We used the most difficult version of the WISDM dataset which is version 1.1 without overlapping time windows
and downloaded it from the UCI machine learning repository.

\end{description}

As can be seen in Table~\ref{DatasetStats}, most of the datasets have a small or medium number of
examples. This is a use case which is just as important as big data since many datasets
that arise in practice have sizes similar to those in the table.  Indeed, most datasets gathered
by TSC are even smaller, but we  did not want to include these since the confidence intervals may become too big.

\begin{table}

\caption{Number of examples, number of timesteps,  number of inputs and number of outputs after one hot encoding.}

\label{DatasetStats}

\begin{tabular}{|c|c|c|c|c|c|c|c|} 
\hline

Dataset &           \#examples & \#timesteps  &  \#inputs &  \#outputs \\
\hline

3W               &  2864       & 64              &  8        &  17 \\  
Crop             & 24000       & 46              &  1        &  24 \\
Double pendulum  & 10000       & 128             &  4        &   4 \\
ECG5000          &  5000       & 140             &  1        &   5 \\
FordB            &  3836       & 500             &  1        &   2 \\
Insect wingbeat  & 20000       &  22             &  200      &  10 \\
LSST             &  4925       &  36             &  6        &  14 \\
WISDM            & 27540       & 40              &  3        &   6 \\
\hline

\end{tabular}

\end{table}

\subsection{Experimental results}
\label{ExperimentalResults}

Table~\ref{ExperimentalComparison} compares an LSTM net with an ARN net for
the test data sets. We used categorical cross entropy (CCE) for the classification
datasets and mean squared error (MSE) for the regression dataset.
The p-value was calculated using McNemar's test for classification and the Wilcoxon signed-rank test
for regression with continuity correction in both cases.
As can be seen in Table~\ref{ExperimentalComparison}, the p-values are better than five sigma for six of the eight datasets.

The column ``Factor better'' shows how many times better the ARN was compared with LSTM with respect to CCE or MSE.

\begin{table}

\caption{Test data cross entropies or MSE and accuracies for LSTM and ARN.}

\label{ExperimentalComparison}

\begin{tabular}{|c|c|c|c|c|c|c|c|}
\hline

Dataset &          LSTM  & ARN  & Factor & LSTM  &  ARN       & p-value\\
        &          CCE/MSE & CCE/MSE &  better &  Acc &   Acc & \\

\hline 

3W               & 0.517  &   0.303 & 1.71  &  0.804  &    0.934  & $1.2  \cdot 10 ^ {-18}$\\
Crop             & 0.796  &   0.661  & 1.20 &   0.742  &    0.775  & $3.8  \cdot 10 ^ {-11}$\\
Pendulum  & 0.154  & 0.110  & 1.40  &   &  &        $1.1  \cdot 10 ^ {-19}$  \\           
ECG5000          & 0.188  &   0.188  &  1 & 0.949  &    0.947  & 0.74 \\
FordB            & 0.510  &   0.140  & 3.64 &  0.759  &    0.950  & $3.6  \cdot 10 ^ {-31}$ \\
Wingbeat  & 1.120  &   1.029  & 1.09 &  0.574  &    0.605  & $1.0  \cdot 10 ^ {-5}$ \\
LSST             & 1.598  &   1.183  & 1.35 & 0.478  &    0.627  & $1.1  \cdot 10 ^ {-21}$ \\
WISDM            & 0.137  &   0.081  & 1.69 & 0.963  &    0.975  & $3.5  \cdot 10 ^ {-8}$ \\

\hline 

\end{tabular}

\end{table}

Since we carefully optimized the hyperparameters for LSTM, the big improvements
obtained for say 3W, Double pendulum,  Ford B, LSST and WISDM, cannot be explained by the LSTMs being poorly tuned.
Instead, the most likely explanation is that the new ARN neurons have a superior modeling and fitting ability.

As can be seen in the table, the improvement that was obtained varies a lot between the datasets.
For example, for the FordB dataset, we obtained a more than three-fold reduction of the test data CCE, whereas there
was no improvement at all for the ECG5000 dataset.
One possible explanation for the huge improvement for FordB is that
this dataset has as many as 500 timesteps and that LSTM cannot effectively handle
so long time dependencies, whereas ARN can do that using skip connections through time.
By looking at the best program for FordB in Appendix~A.4, it is easy to see that
skip connections are indeed used, but otherwise this is a really hard neuron to analyze
since it was not designed by human beings.

Since the experiments optimized cross entropy instead of accuracy, it is likely that somewhat  higher accuracies could be obtained
if ADATE were to directly optimize the accuracy instead.

We omitted the neuron for ECG5000 since it was no improvement, 
but the best ones for all the other datasets are listed in the appendices.
As can be seen there, there is a huge variation from one dataset to the next, but both skip connections and
the linear combinations of state 0 ({\tt OtherPeepsLC }) are frequently used.
From the outset, we naively guessed that reading the memories of other neurons would give rise to problems akin
to vanishing or exploding gradients, but this seems to be much less problematic than we thought.

All of the best programs use novel activation functions that we have never seen before.
For example, the best program for WISDM in Appendix~A.7 uses {\tt srelu} applied
to quadratic forms in several places. Many of the other neurons also contain various
quadratic expressions, but which roles they have is not clear.

There are some obvious redundancies in the coding, for example applying {\tt relu} twice.
ADATE is able to remove such redundancies, but much more run time might be needed.

In general, evolution in ADATE is so effective that neurons are likely to become exquisitely
adapted to the special characteristics of the dataset. This means that neurons should not
be reused from one class of datasets to the next, that is, a new ADATE run is needed for each new type of application.
One way to avoid this special adaption would be to run on say hundreds of different datasets  simultaneously
using many more cores and possibly accelerators.

\subsection{A closer look at some ARNs}

In this section, we will discuss the circuit diagrams for the best LSST neuron and the best FordB neuron.
These were selected since they are among the simpler best neurons.
In order to translate the neuron syntax from SML to diagrams, we first printed them as C expressions using our compiler,
converted them to equations and then manually simplified these equations before using them to draw the neurons.

\begin{figure}[!htbp]
\begin{center}
\scalebox{0.30}
 {
   {\includegraphics{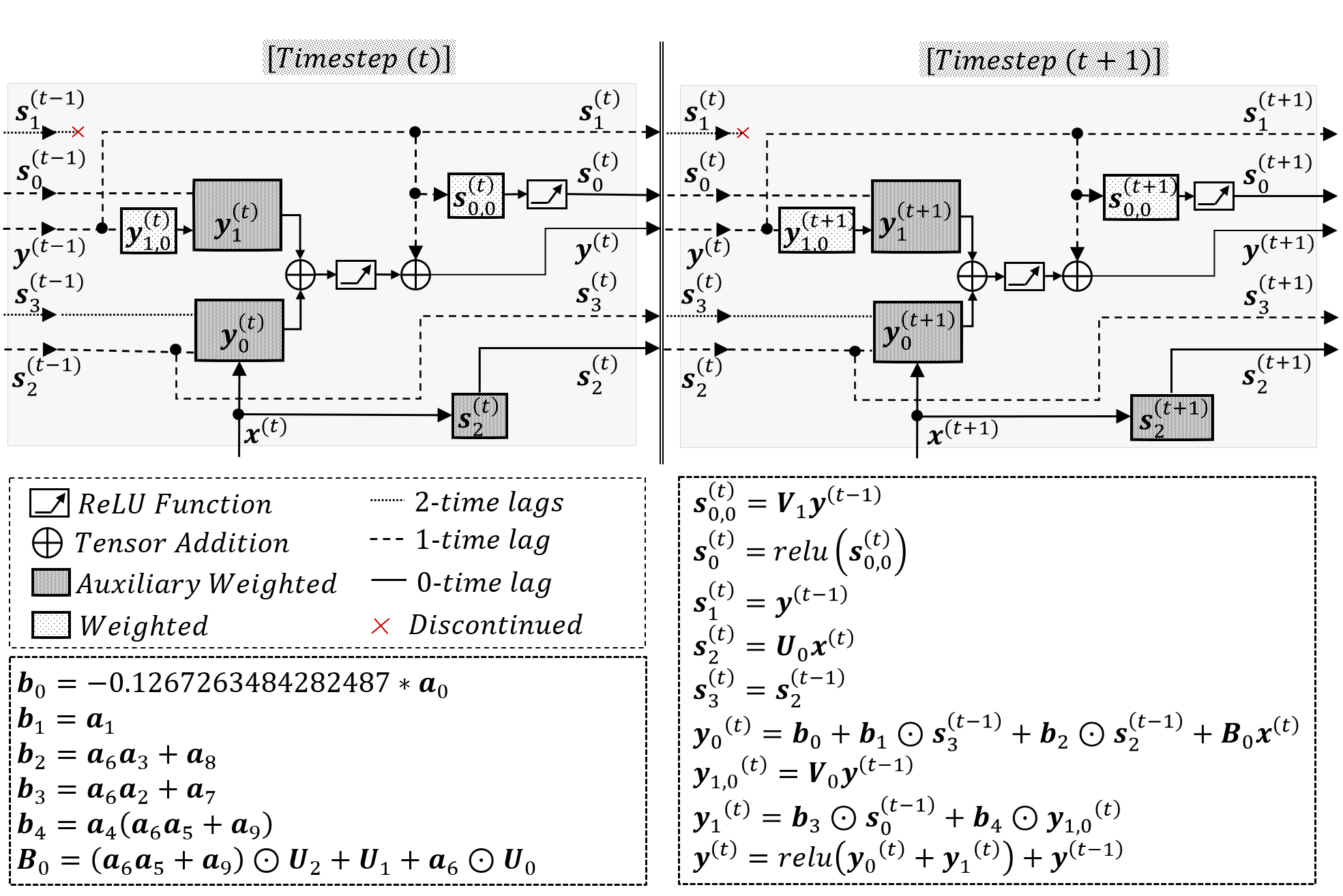}}
 }
\end{center}
\caption{A circuit diagram for the best neuron for the LSST dataset .}
\label{LSSTFig}
\end{figure}

Both the diagram and the equations for LSST are shown in Figure~\ref{LSSTFig}.
All vector-vector multiplications are element-wise  and
all variables denoted by {\bf b} or {\bf B} are vectors or matrices manually derived from
{\tt cons} expressions in the SML code that after translation to C contains auxiliary weights, denoted
by {\bf a} variables. We do not show the full and lengthy manual translation  from SML to simplified equations.

It is a bit fascinating to see that ADATE actually has invented a small convolutional filter represented
by the equation for ${\bf y_0^{(t)}}$.  This filter combines the inputs from three consectutive timesteps
using a linear combination. However, the output from the filter is combined with skip connections from
previous outputs represented by ${\bf y_1^{(t)}}$.
Thus, the convolution is in principle over inputs and some outputs and the 
result is  fed to {\tt relu} as can be seen in the equation for  ${\bf y^{(t)}}$.

Another interesting feature of this neuron is that it totally has abandoned {\tt tanh} and {\tt srelu} as
activation functions. As can be seen in  Figure~\ref{LSSTFig}, it only contains two {\tt relu} functions.
However, we are not able to theoretically understand the  dynamic behaviour of RNNs that contain such neurons
and can only claim that it produces superior accuracy for the LSST dataset.

\begin{figure}[!htbp]
\begin{center}
\scalebox{0.35}
 {
   {\includegraphics{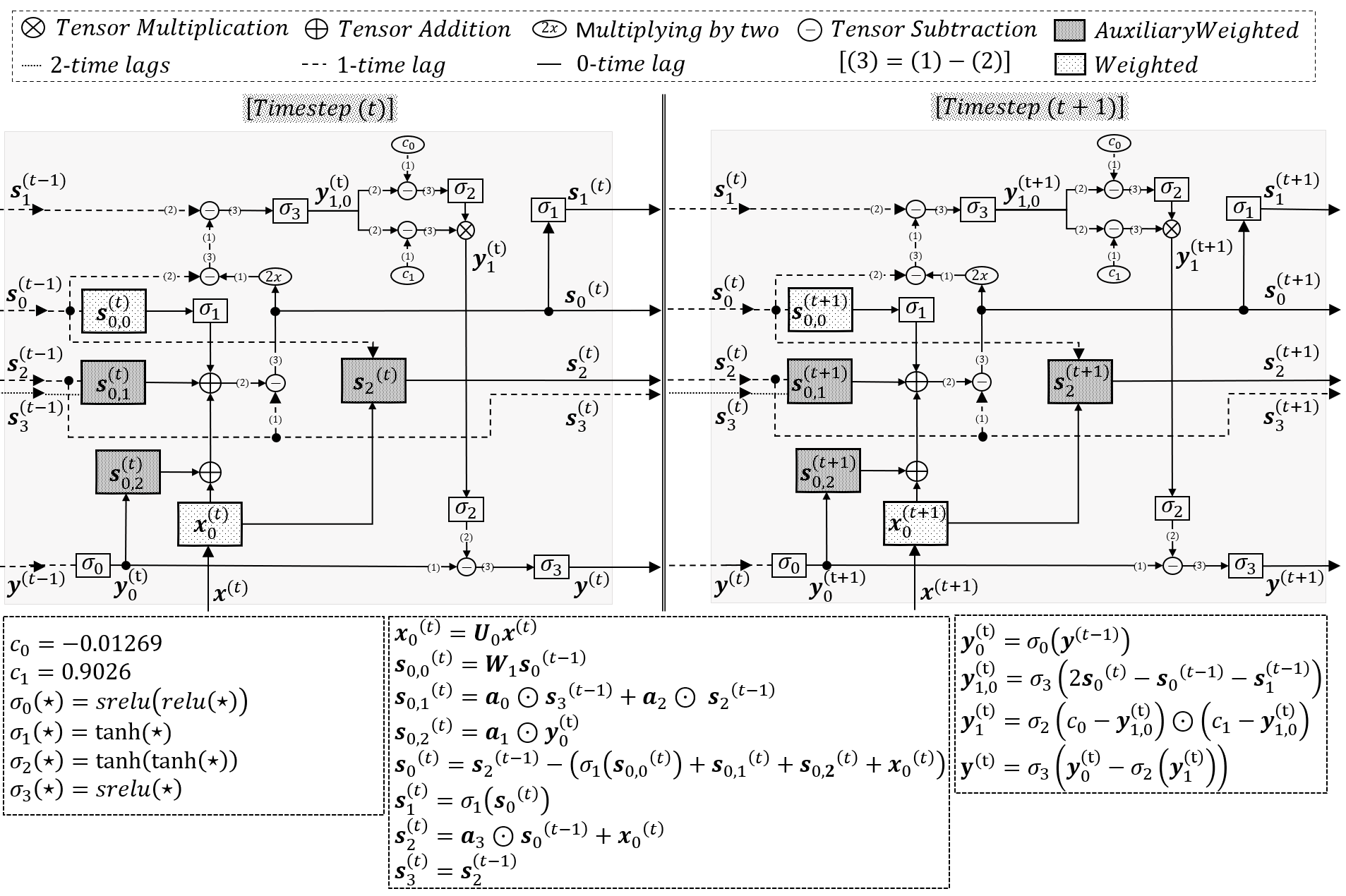}}
 }
\end{center}
\caption{A circuit diagram for the best neuron for the FordB dataset .}
\label{FordBFig}
\end{figure}

The best neuron for the FordB dataset is more complex and shown in Figure~\ref{FordBFig}.
We will now take a look at some elaborate skip connections in this neuron that use
the output from time $t-4$ to calculate the output at time $t$.
Since the dataset contains 500 timesteps, we can speculate that it is especially important
to curate information over long time periods and that special connections
may have been evolved to handle that.

Let $\rightarrow$ mean that the left hand side is partially used to compute the right hand side.
Then, the information flow from $y^{(t-4)}$ to  $y^{(t)}$ is as follows
as can be seen in the circuit diagram.

\[
y^{(t-4)} \rightarrow s_0^{(t-3)}  \rightarrow s_2^{(t-2)} \rightarrow s_3^{(t-1)} \rightarrow y^{(t)}
\]

Thus, the internal states are employed to implement information flow that appears to be custom designed
to retain long term memory for the FordB dataset.

As a curiosity, we can just mention that ADATE has discovered ARNs that retain long term memory much better than
LSTM for thousands of timesteps when simulating  complex Mealy machines.

\subsection{Analysis of a Pareto front}
\label{ParetoFront}

\begin{figure}[!htbp]
\begin{center}
\scalebox{0.32}
 {
   {\includegraphics{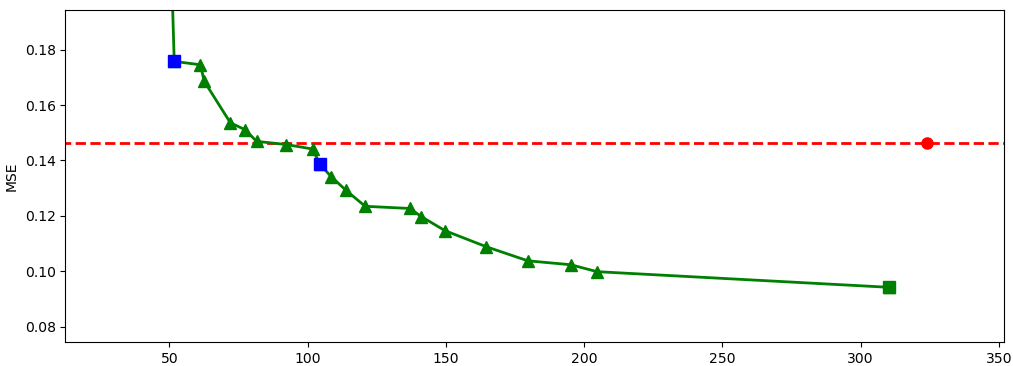}}
 }
\end{center}
\caption{The Pareto front for Double pendulum.}
\label{ParetoFrontFig}
\end{figure}

Recall that a Pareto front contains gradually bigger and better programs so that the
smallest and worst program comes first whereas the last program is the biggest and the best.
As an example, we will use the Pareto front for the Double pendulum dataset
which consists of 24 programs. The first four of these are so small
that they are not interesting but the last 20 are shown in Figure~\ref{ParetoFrontFig}.
For this dataset, an LSTM has a validation MSE of 0.146 as indicated by the horizontal line
in Figure~\ref{ParetoFrontFig}, which means that all programs below the line are better than LSTM
for Double pendulum on validation data.

By using Pareto fronts, ADATE tries to both minimize MSE and program size.
This has been so successful that all programs in the front are smaller, at least according to the ADATE syntactic complexity,
 than the LSTM program, which has
a size of 324 bits. Thus, ADATE has found a plethora of programs that are both smaller and better than LSTM
for Double pendulum.

We will take a closer look at the two programs that are indicated by the blue squares in  Figure~\ref{ParetoFrontFig}.
The smallest one, corresponding to the leftmost square, has  
an MSE of 0.176 and is the following simple RNN, where no internal neuron states are used.
Therefore, we have also manually edited the program  to remove them 

\begin{verbatim}
relu( lc2( cons( lc1( OtherOutputsLC ), InputsLC ) ) )
\end{verbatim}

This is easy to understand and just a linear combination of the inputs and the outputs from other neurons, albeit
formulated as a linear combination of a linear combination. Apparently, {\tt relu} is 
more suitable than the other activation functions.

The next square square in  Figure~\ref{ParetoFrontFig} is the program in Figure~\ref{PendulumProgram}
that has an MSE of 0.139, which is
better than for the LSTM. 

Even if this neuron is quite simple, it is already becoming  hard to understand
exactly why it is better, with analysis being impeded by the use
of internal states from other neurons.  A partial hypothesis may be
that using quadratic base functions may give more suitable regression
modeling of a double pendulum.

The best neuron for Double Pendulum, which is the green and last square in  Figure~\ref{ParetoFrontFig},
is given in Appendix~A.3.
It is much better  and has an 
MSE of 0.094 on validation data. It is also much more
difficult to understand. 

Also, note that the test data MSE for this neuron is 0.11 and that
using validation data for both hyperparameter optimization, ADATE runs, and model checkpointing leads
to some overfitting on validation data. In spite of this,
the neuron is significantly better than LSTM also on test data 
with a p-value around $10^{-19}.$

The syntactic complexity makes a big jump with little gain in MSE
from the second last to the best program in the front.
Due to Ockhams's razor \citep{Ockham}, overfitting can be suspected when this happens.

\section{Conclusions and future work}

We tested the ADATE recurrent neurons (ARNs) on eight datasets.
For the four best datasets, the ARNs were a factor of 3.6, 1.7, 1.7 and 1.4 better respectively
as measured by cross entropy or mean squared error on test data. The p-values for these four were
below about $10^{-18}$, which is clearly better than five sigma.

For the other four datasets, there were significant improvements for  three but failure for one.
The reason it did not work for one dataset seems to be overfitting to the validation data caused by a combination
of hyperparameter tuning, model check pointing and using validation  data to calculate
evaluation values for programs generated by ADATE and selecting the best one.

The results are so good that ARNs should be tried in practice whenever recurrent nets are used.
The reasons for the big improvements that were found for at least half of the datasets are as follows.

\begin{enumerate}

\item ADATE is quite capable when it comes to automatically adapting the code of a neuron to a specific class of datasets.

\item The ARNs contain several somewhat unusual ingredients such as
\begin{enumerate}

\item Up to four internal states per neuron. 
Sometimes, these are apparently  used by ADATE to construct intricate skip connections through time.

\item Linear combinations of the states of other neurons in addition to the usual combinations of outputs.

\item Custom made activation functions that sometimes contain quadratic forms.

\item Differentiation between ``self'' and ``all others''.

\end{enumerate}
\end{enumerate}

The technology is ready to use right away in {\CC } with an option to manually port to Tensorflow.
We encourage people interested in using it to contact 
the corresponding author of this paper. However, a cluster with between 500 and 1000 cores is needed to run on
a subset of data with up to about 50 000 training examples for about three days. 
Thus, the number of datasets that can be processed
may be somewhat limited by available computational resources.

Here are some possibilities for future work.

\begin{enumerate}

\item Let an ARN layer be an ensemble of different neuron types and use ADATE to evolve
all the different types, either one type at a time or all types simultaneously.

\item Use many stacked ARN layers with different neuron types in each layer.

\item Find a way to efficiently optimize hyperparameters as a part of the evolution
instead of doing it only once at the beginning. 

\item Run on hundreds of datasets during the same evaluation in order to find neurons
that are generally better. However, we expect that
evolving neurons for specific classes of datasets often will yield much better results.

\item Study other neural architectures than RNNs. For example, the  batch normalization and ResNet blocks
in a CNN could be viewed as the function {\tt f} to improve and the time dimension in an RNN could be
replaced by a layer index dimension, where say four states per block  are curated  from one layer to the next
by {\tt f}. 

\item Another possibility would be define differentiable
functional programs and neural nets that represent them. These could use differentiable datastructures
and handle recursion using a predefined set of differentiable schemas that are generalizations of say catamorphisms, 
anamorphisms and hylomorphisms.

\end{enumerate}

It may also be possible to employ the automatic programming presented in this paper for a variety of  other 
current and future neural net and differentiable programming technologies.

\appendix

\section*{Appendix A. The best neurons according to validation data}

This appendix lists the best programs according to validation data for the seven  datasets out of eight for
which improvements were found. They are presented exactly as they were printed by the ADATE system and could
typically be somewhat simplified manually. Thus, they may look more complex than they really are.

The best neuron for WISDM has also been implemented by one of us as a custom layer in Tensorflow in addition to
the automatically generated {\CC } code that our compiler generates based on
the SML programs. 

Requests for  neuron synthesis and {\CC } or Tensorflow implementations  
for other datasets are very welcome.

\subsection*{Appendix A.1. The best ARN for 3W}              
\begin{verbatim}
fun f
      (
        SelfPeep0,
        SelfPeep1,
        SelfPeep2,
        SelfPeep3,
        SelfOutput,
        OtherPeepsLC,
        OtherOutputsLC,
        InputsLC
        ) = 
  case cons( SelfOutput, cons( SelfOutput, InputsLC ) ) of
    V21902689 =>
  case
    (
      SelfOutput,
      tanh(
        srelu(
          (
            (
              lc3( cons( lc1( V21902689 ), V21902689 ) ) *
              srelu( SelfPeep3 )
              ) -
            (
              tanh(
                lc0(
                  cons(
                    SelfPeep3,
                    cons( tanh( SelfOutput ), InputsLC )
                    )
                  )
                ) +
              srelu(
                relu(
                  tanh(
                    lc2(
                      cons(
                        ~0.14531347527330391E~1,
                        cons(
                          SelfPeep3,
                          cons(
                            srelu( srelu( SelfPeep2 ) ),
                            cons(
                              SelfOutput,
                              cons( SelfPeep0, InputsLC )
                              )
                            )
                          )
                        )
                      )
                    )
                  )
                )
              )
            )
          )
        )
      )   of
    ( V21915A57, V21915A58 ) =>
      (
        V21915A58,
        SelfPeep0,
        srelu( relu( lc2( OtherOutputsLC ) ) ),
        SelfPeep0,
        srelu(
          (
            V21915A57 -
            relu(
              tanh(
                tanh(
                  srelu(
                    srelu(
                      case ( SelfPeep1, V21915A58 ) of
                        ( V21915A59, V21915A5A ) =>
                          ( V21915A5A - V21915A59 )
                      )
                    )
                  )
                )
              )
            )
          )
        )
\end{verbatim}

\subsection*{ Appendix A.2. The best ARN for Crop}           
\begin{verbatim}
fun f
      (
        SelfPeep0,
        SelfPeep1,
        SelfPeep2,
        SelfPeep3,
        SelfOutput,
        OtherPeepsLC,
        OtherOutputsLC,
        InputsLC
        ) = 
  case
    (
      SelfOutput,
      (
        ( lc3( OtherPeepsLC ) * lc0( InputsLC ) ) -
        (
          tanh(
            srelu(
              lc0(
                cons(
                  SelfPeep1,
                  cons(
                    SelfPeep3,
                    cons( lc4( OtherPeepsLC ), bias )
                    )
                  )
                )
              )
            ) +
          lc3( InputsLC )
          )
        )
      )   of
    ( VDBC64A, VDBC64B ) =>
  case srelu( srelu( ( VDBC64B - SelfPeep1 ) ) ) of
    V1C427005 =>
      (
        VDBC64B,
        SelfPeep0,
        lc0( cons( SelfPeep0, OtherPeepsLC ) ),
        SelfPeep2,
        srelu(
          case ( V1C427005, V1C427005 ) of
            ( V1C481BD5, V1C481BD6 ) =>
          let
            fun g1CDF7C46 V1CDF7C47 =
              ( VDBC64A - V1CDF7C47 )
          in
            g1CDF7C46( ( V1C481BD6 * V1C481BD5 ) )
          end 
          )
        )
\end{verbatim}

\subsection*{Appendix A.3. The best ARN for Double pendulum}
\label{DoublePendulumBest}
\begin{verbatim}
fun f
      (
        SelfPeep0,
        SelfPeep1,
        SelfPeep2,
        SelfPeep3,
        SelfOutput,
        OtherPeepsLC,
        OtherOutputsLC,
        InputsLC
        ) = 
  (
    (
      relu( relu( lc1( cons( lc1( OtherPeepsLC ), InputsLC ) ) ) ) -
      relu(
        lc3(
          cons(
            lc4( OtherPeepsLC ),
            cons( 0.22174599383632232, InputsLC )
            )
          )
        )
      ),
    SelfPeep1,
    SelfPeep2,
    SelfPeep3,
    (
      (
        (
          lc4(
            cons(
              lc4( OtherPeepsLC ),
              cons(
                lc0(
                  cons(
                    SelfPeep2,
                    cons(
                      ( ~0.6961519103176064 + SelfPeep1 ),
                      InputsLC
                      )
                    )
                  ),
                InputsLC
                )
              )
            ) *
          lc0( OtherPeepsLC )
          ) +
        SelfPeep0
        ) *
      (
        SelfPeep1 -
        (
          tanh(
            case
              cons(
                lc3(
                  cons(
                    SelfPeep2,
                    cons( ~0.6961519103176064, InputsLC )
                    )
                  ),
                InputsLC
                )             of
              V1662138D =>
                (
                  srelu(
                    lc0(
                      cons(
                        SelfPeep0,
                        cons( lc0( V1662138D ), InputsLC )
                        )
                      )
                    ) *
                  srelu(
                    lc0(
                      cons(
                        relu(
                          lc3(
                            cons(
                              SelfPeep0,
                              cons(
                                lc0(
                                  cons( lc0( V1662138D ), V1662138D )
                                  ),
                                InputsLC
                                )
                              )
                            )
                          ),
                        InputsLC
                        )
                      )
                    )
                  )
            ) +
          ~0.45284110969509517
          )
        )
      )
    )
\end{verbatim}

\subsection*{Appendix A.4. The best ARN for FordB}         
\label{FordBBest}
\begin{verbatim}
fun f
      (
        SelfPeep0,
        SelfPeep1,
        SelfPeep2,
        SelfPeep3,
        SelfOutput,
        OtherPeepsLC,
        OtherOutputsLC,
        InputsLC
        ) = 
  case srelu( srelu( relu( relu( SelfOutput ) ) ) ) of
    V281AF3E7 =>
  case
    (
      V281AF3E7,
      (
        SelfPeep2 -
        (
          tanh( lc1( OtherPeepsLC ) ) +
          lc0(
            cons(
              SelfPeep3,
              cons( V281AF3E7, cons( SelfPeep2, InputsLC ) )
              )
            )
          )
        )
      )   of
    ( VDBC64A, VDBC64B ) =>
      (
        VDBC64B,
        tanh( srelu( srelu( VDBC64B ) ) ),
        lc0( cons( tanh( SelfPeep0 ), InputsLC ) ),
        SelfPeep2,
        srelu(
          (
            VDBC64A -
(             case
              case
                let
                  fun g1C57E5DF V1C57E5E0 =
                    (
                      V1C57E5E0,
                      (
                        ~0.12690104588539253E~1 -
(                         case ( VDBC64B, SelfPeep1 ) of
                          ( V1C57EDC3, V1C57EDC4 ) =>
                            srelu(
                              (
                                ( V1C57EDC3 - SelfPeep0 ) +
                                ( V1C57EDC3 - V1C57EDC4 )
                                )
                              ) )
                        )
                      )
                in
                  g1C57E5DF(
                    case g1C57E5DF 0.1029527350644156 of 
                      ( A, B) => A + B
                    )
                end                of
                ( V5FAAD0B, V5FAAD0C ) =>
                  ( V5FAAD0B, srelu( V5FAAD0C ) )             of
              ( V1D190879, V1D19087A ) =>
                tanh( tanh( ( V1D19087A * V1D190879 ) ) ) )
            )
          )
        )
\end{verbatim}

\subsection*{Appendix A.5. The best ARN for Insect wingbeat}
\begin{verbatim}
fun f
      (
        SelfPeep0,
        SelfPeep1,
        SelfPeep2,
        SelfPeep3,
        SelfOutput,
        OtherPeepsLC,
        OtherOutputsLC,
        InputsLC
        ) = 
  case
    (
      SelfOutput,
      (
        (
          srelu(
            tanh(
              tanh(
                lc3(
                  cons(
                    0.10022791851659179E1,
                    cons( lc4( OtherOutputsLC ), InputsLC )
                    )
                  )
                )
              )
            ) *
          srelu( tanh( lc0( InputsLC ) ) )
          ) -
        (
          tanh( srelu( ~0.6873893912995532E~2 ) ) +
          srelu( relu( SelfOutput ) )
          )
        )
      )   of
    ( VDBC64A, VDBC64B ) =>
  case tanh( srelu( srelu( tanh( relu( VDBC64B ) ) ) ) ) of
    V1C9C2AC6 =>
  case ( V1C9C2AC6, V1C9C2AC6 ) of
    ( V1C9C2AD2, V1C9C2AD3 ) =>
      (
        VDBC64B,
        ( tanh( SelfPeep2 ) * SelfPeep3 ),
        relu( tanh( relu( V1C9C2AD3 ) ) ),
        SelfPeep2,
        srelu( ( VDBC64A - tanh( V1C9C2AD2 ) ) )
        )
\end{verbatim}

\subsection*{Appendix A.6. The best ARN for LSST}          
\label{LSSTBest}
\begin{verbatim}
fun f
      (
        SelfPeep0,
        SelfPeep1,
        SelfPeep2,
        SelfPeep3,
        SelfOutput,
        OtherPeepsLC,
        OtherOutputsLC,
        InputsLC
        ) = 
  (
    relu( lc1( OtherOutputsLC ) ),
    SelfOutput,
    lc0( InputsLC ),
    SelfPeep2,
    (
      relu(
        lc1(
          cons(
            srelu( ~0.1267263484282487 ),
            cons(
              SelfPeep3,
              case
                cons(
                  SelfPeep0,
                  cons(
                    SelfPeep2,
                    cons(
                      lc2( cons( lc0( OtherOutputsLC ), InputsLC ) ),
                      InputsLC
                      )
                    )
                  )               of
                V259F26F2 => cons( lc0( V259F26F2 ), V259F26F2 )
              )
            )
          )
        ) +
      SelfOutput
      )
    )
\end{verbatim}

\subsection*{Appendix A.7. The best ARN for WISDM}        
\label{WISDMBest}
\begin{verbatim}
fun f
      (
        SelfPeep0,
        SelfPeep1,
        SelfPeep2,
        SelfPeep3,
        SelfOutput,
        OtherPeepsLC,
        OtherOutputsLC,
        InputsLC
        ) = 
  case
    (
      SelfOutput,
      (
        lc0( bias ) -
        (
          tanh(
            srelu(
              srelu(
                lc1(
                  cons(
                    SelfPeep1,
                    cons(
                      SelfPeep3,
                      cons(
                        (
                          lc4( InputsLC ) *
                          lc2( cons( SelfPeep0, InputsLC ) )
                          ),
                        cons(
                          srelu(
                            relu(
                              lc2(
                                cons(
                                  lc0( OtherOutputsLC ),
                                  OtherPeepsLC
                                  )
                                )
                              )
                            ),
                          case
                            cons( lc1( OtherPeepsLC ), InputsLC ) 
                          of
                            V271482D0 =>
                              cons( lc2( V271482D0 ), V271482D0 )
                          )
                        )
                      )
                    )
                  )
                )
              )
            ) +
          srelu(
            relu( lc2( cons( lc4( OtherPeepsLC ), InputsLC ) ) )
            )
          )
        )
      )   of
    ( VDBC64A, VDBC64B ) =>
      (
        VDBC64B,
        SelfPeep0,
        srelu( srelu( srelu( lc0( OtherPeepsLC ) ) ) ),
        SelfPeep2,
        srelu(
          (
            VDBC64A -
            (
              ( VDBC64B - SelfPeep1 ) *
              ( VDBC64B - SelfPeep1 )
              )
            )
          )
        )
\end{verbatim}

\vskip 0.2in
\bibliography{arn4}

\end{document}